%% file: root.tex
\documentclass[letterpaper, 10 pt, conference]{ieeeconf}  % Comment this line out if you need a4paper

\IEEEoverridecommandlockouts                              % This command is only needed if 
\title{\LARGE \bf Making Sense of Touch from the Child’s View for Contrastive Learning}

\input{preamble}

\input{authors}
\begin{document}

\maketitle
\thispagestyle{fancy}
% \thispagestyle{empty}

%%%%%%%%%%%%%%%%%%%%%%%%%%%%%%%%%%%%%%%%%%%%%%%%%%%%%%%%%%%%%%%%%%%%%%%%%%%%%%%%
\input{abstract}

\pagestyle{empty}

%%%%%%%%%%%%%%%%%%%%%%%%%%%%%%%%%%%%%%%%%%%%%%%%%%%%%%%%%%%%%%%%%%%%%%%%%%%%%%%%
\input{intro}
\input{relatedworks}
\input{dataset}

\input{experiments}

\input{conclusions}

\input{appendix}
\input{acknowledgemnts}

\bibliographystyle{IEEEtran}
\bibliography{bibliography}

\end{document}

%% file: preamble.tex
\usepackage{xcolor}

\usepackage{graphicx}
\usepackage{multirow}
\usepackage{tabularx}
\usepackage{longtable}
\usepackage{cite}

\usepackage{subcaption}

\usepackage{pifont}
\usepackage{booktabs}
\usepackage{colortbl}
\usepackage{makecell}

\makeatletter
\let\NAT@parse\undefined
\makeatother
\usepackage{listings}
\usepackage{hyperref}
\definecolor{groupA}{RGB}{248,250,253}  % blue grey
\definecolor{groupB}{RGB}{252,249,245}  % light yellow
\definecolor{groupC}{RGB}{247,250,247}  % green grep
\definecolor{groupD}{RGB}{250,247,250}  % pink purple
\definecolor{groupE}{RGB}{240,240,240}  % very light gray

\newcommand{\eat}[1]{}

\usepackage{fancyhdr}

%% file: authors.tex
\author{Max Whitton, Zecheng Wang$^\star$, Puchen Liu$^\star$, Quang Tuan Truong, Shengao Wang, Manaswi Yadamreddy, \\
Oktay Ozel, Visista Jayanti, Saniya Sekhon, Hanna Samuel Tadesse, Lawrence Miao, Junjie Wang, \\
Jiasen Lu, Chen Yu$^1$, and Boqing Gong% <-this % stops a space
\vspace{10pt}\\
\textbf{Project Homepage:} \url{https://max-whitton.github.io/Contrastive_Learning_From_Touch}%\href{https://max-whitton.github.io/Contrastive_Learning_From_Touch}{\textbf{Project Homepage}} 
\thanks{Boston University,
        {\tt\small \{maxwh,bgong\}@bu.edu}}%
\thanks{$^{1}$University of Texas, Austin,        {\tt\small chen.yu@austin.utexas.edu}}% 
}

%% file: abstract.tex
\begin{abstract}
Is the sense of touch a mechanism for human babies' learning of visual concepts? If so, can we quantify its importance, and to what extent do babies rely on their sense of touch for visual learning? To approach these questions in a principled way, we propose a structured coding system for baby-centric touch events, yielding a dataset of 264k two-second clips of touch events coded according to this system. Using this dataset, we pretrain developmentally grounded models that reveal promising insights into the nature of baby learning from touch.\\
\end{abstract}

%% file: intro.tex
\section{INTRODUCTION}

Developmental psychology has a long history of linking motor skills and physical interactions with babies' perceptual abilities, tracing back to the foundational theories of Piaget~\cite{piaget1952origins}, Gibson~\cite{gibson2014ecological}, and Thelen \& Smith~\cite{thelen1994dynamic}. For instance, developmental changes in motor skills, such as sitting and manual exploration, contribute directly to three-dimensional object completion~\cite{soska2010systems}. Furthermore, self-generated variability in object images---achieved through holding, moving, or stacking objects---strongly predicts vocabulary growth~\cite{slone2019self}. These studies suggest that reentrant mappings~\cite{gershkoff2005building,edelman2013reentry,six-lessons}, in which activity in the visual system is correlated in real time with that of the haptic system, provide qualitatively different glosses on the world that educate each other and facilitate robust conceptual learning. 

We present a computational framework that investigates the role of touch in babies' acquisition of visual concepts, asking whether, where, and to what extent touch signals improve visual learning.

We build the work upon recent computational efforts to model babies' audiovisual intake~\cite{vongGroundedLanguageAcquisition2024,wangBabyVLMDataEfficientPretraining2025,wangbabyvlmv22026}. Notably, the child's view for contrastive learning (CVCL)~\cite{vongGroundedLanguageAcquisition2024} successfully demonstrated how a generic neural network can learn representations by mapping transcribed spoken language to corresponding raw, egocentric visual streams. While this audiovisual paradigm provides a strong foundation for modeling a baby's language acquisition, by relying exclusively on seeing and hearing, such models omit the physical interactions (e.g., grasps, pokes, and manipulations) that babies actively use to disambiguate, reinforce, and educate their visual systems.

Ideally, our framework will mirror a baby's sensory experience as closely as possible, recovering as much haptic input as we can from their real life, without adding any extra data that a baby would not have access to. However, it is prohibitively difficult to capture a baby's high-fidelity tactile sensations at scale directly. We approach this problem in a principled way, without introducing intrusive physical sensors, by leveraging a structured coding system for babies' touch events. We apply this taxonomy of touch to longitudinal, egocentric video recordings from three infants' perspectives, yielding a novel dataset of 264k carefully coded video clips of touch events. This dataset effectively captures tactile experience by discretizing visual evidence of intentional physical interactions by or with the baby. 

With this dataset, we extend CVCL by contrasting time-locked visual-speech and visual-touch pairs with time-misaligned ones. We then evaluate the learned representations on existing approximations of baby perception tests. Our findings indicate that touch is a highly effective learning signal, particularly when data from the audiovisual modalities is limited---the sense of touch enables the model to achieve superior linear probing and zero-shot performance on visual recognition tasks compared to baselines that learn solely from visual-speech pairs. These findings provide promising insights into the data-efficient nature of baby learning. Our data processing, training, and evaluation code can be found, along with out full dataset and some of our pretrained model weights, on our  \href{https://max-whitton.github.io/Contrastive_Learning_From_Touch/}{project homepage}. Together, these serve as a computational sandbox to study related theories of development and learning.

%% file: dataset.tex
\section{Coding Touch}

% \boqing{Answer why we need to manually label TOUCH. What are the pros and cons compared to measuring TOUCH from hardware sensors? From those we can briefly describe the overarching methodology, followed by the detailed section below.}

%\jiasen{Agree with boqing's comment, current section didn't answer the motivation to annotate touch from video instead of directly using hardware sensors. the scale can be the reason. then, should answer why need to manually label touch. It will be good to have fig 1 or detailed examples how the dataset looks like.}

We describe our approach to coding touch events from babies' headcam videos. Figure~\ref{fig:data-pipeline} illustrates this pipeline.

\subsection{Overview and Design Considerations}
\subsubsection{Data source}
To align our data with what a real baby would have access to, we use SAYCam~\cite{sullivanSAYCamLargeLongitudinal2021}, a corpus of longitudinal, egocentric 478-hours recordings from three infants taken once every week from roughly 6 to 32 months old. While similar data sources exist, such as BabyView~\cite{longBabyViewDatasetHighresolution2025}, we choose SAYCam because of the availability of existing in-domain benchmarks, pretraining splits, and comparable models~\cite{vongGroundedLanguageAcquisition2024, wangbabyvlmv22026}. In fact, we use the exact same $[\texttt{video frame, speech transcript}]$ training pairs as BabyVLM-V2 and supplement them with our $[\texttt{vision, touch}]$ learning events.

\subsubsection{Coding vs.\ physical sensors for touch}
Section~\ref{sec:touch-taxoonmy} explains the coding system for touch from babies' headcam videos --- how we define and code touch events specific to babies. While collecting touch sensations using physical sensors initially seems elegant, we choose to code touch events manually because a) the scale of paired $[\texttt{vision, sensor}]$ data required for training a computation model would be prohibitively expensive to acquire using physical sensors, and b) we hypothesize the mechanism behind babies' touch learning to be more discrete and categorical than continuous physical measurements provide. 

\subsubsection{Coding touch events at scale}
To obtain a large-scale dataset of such events (Section~\ref{sec:label-at-scale}), we develop a pipeline that semi-automatically annotates video clips using our baby touch taxonomy, yielding a text caption for each touch event. Examples from the completed dataset are shown in Figure \ref{fig:models}A. 

\subsubsection{Mapping text descriptions of touch to cluster indices}
While the text descriptions of touch events facilitate coding and discussion among raters, they are semantic rather than tactile. Training a computational model on generated text descriptions would give more semantic supervision than a baby would receive. Hence, we discretize the ``touch captions'' to discrete cluster indices, each cluster index corresponding to a unique type of physical experience. 

Altogether, our approach aims to construct the most accurate possible reconstruction of baby touch learning to reliably explore the role of touch in developmental learning.

\begin{figure}
    \centering
    \includegraphics[width=\linewidth]{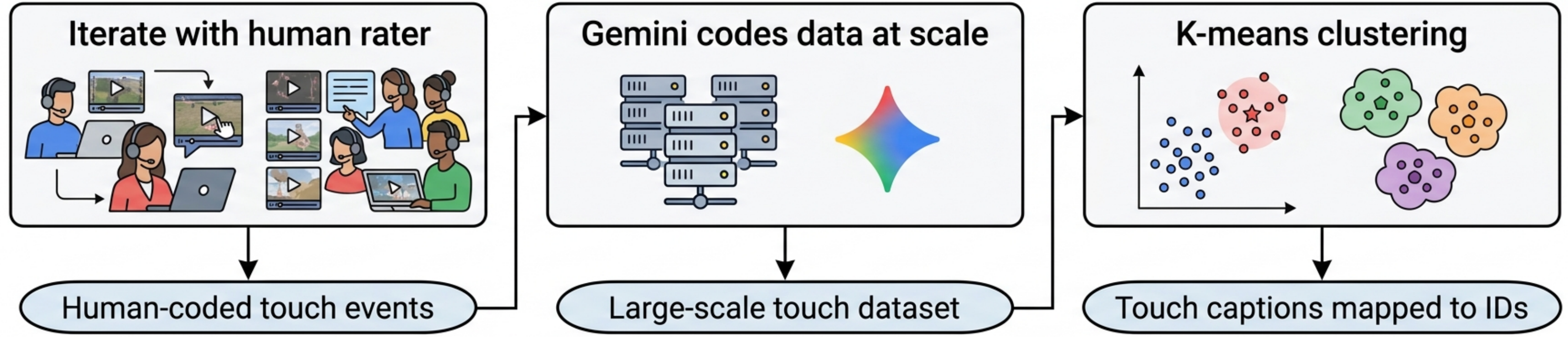}
    \caption{Three-stage pipeline to code a large scale dataset of baby-centric touch events.}
    \label{fig:data-pipeline}
\end{figure}

% \subsection{Existing audio labels}
% We supplement our proposed dataset with \cite{wangbabyvlmv22026}'s (vision, audio) pretraining pairs, believing that the more senses we attempt to recover in our training, the closer we can mimic the way a human baby learns.

\subsection{Human-coding of touch (Figure~\ref{fig:data-pipeline}, stage 1)} \label{sec:touch-taxoonmy}
To model learning from touch sensations, we need a structured touch taxonomy that is simultaneously consistent, such that the same touch event is generally mapped to a single code, and detailed, such that the set of codes contain sufficient granularity to describe the variations that a baby experiences between touch events. To help design such a taxonomy, we begin by recruiting a team of fifteen university participants. The importance of human annotators is two-fold: First, we need reliable, intelligent annotators who are sensitive to edge cases to design a robust structure for the taxonomy in the first place. Second, we need a set of trusted human annotations as a validation check for the quality of the semi-automatically generated event detections of touch, detailed later. 
% Our taxonomy attempts to satisfy these requirements and can be seen in \todo{FIGURE OF BREAKDOWN MADE BY TUAN}.  \\

\subsubsection{Iterative design process}
We met with our raters in person weekly to design a consistent, detailed partition of a baby's touch experience. They first coded a small set of video frames using a basic handcrafted taxonomy. From their results, we compute frequency and inter-rater agreement statistics for each category. Based on common points of ambiguity or lack of detail, we updated the taxonomy by redefining or splitting unclear categories. For example, the categories ``holding'' and ``manipulating'' had high inter-rater uncertainty, so we merged them into ``holding/manipulating'', trading expressiveness for consistency. Conversely, ``lightly touching" initially represented too many different kinds of touch sensations, so we added a new label ``poking" to increase the early taxonomy's expressiveness. We repeated this process three times on fresh sets of video clips until achieving satisfactory label distributions (shown in Figure \ref{fig:piecharts}) and inter-rater agreement, then moved on to a scaled labeling effort.

\subsubsection{Inclusion vs.\ Exclusion}
We define an infant touch event as a discrete, intentional action done by or to a baby. Static touch sensations (e.g., the sensation of a chair on one's butt) and observational touch (e.g., the baby watching another person perform) are not classified as touch events in our framework. However, if an outside agent (usually a parent) intentionally touches the baby's body, such as with a rub or pat, we consider it a learnable touch event. Any time the person performing or receiving an action, or the action itself, is not immediately obvious, we reject it from our dataset. 

\subsubsection{Taxonomy structure}
From the results of our iterative design process, we find that we can effectively parameterize a touch event with its \textit{\textcolor{red}{verb}, \textcolor{orange}{noun}, \textcolor{green!60!black}{body part}}, and \textit{\textcolor{blue}{additional conditions}}. For example, a touch event could be classified as [\textcolor{red}{manipulating}, \textcolor{orange}{spoon}, \textcolor{green!60!black}{hand}, \textcolor{blue}{eating setting}]. Note that \textcolor{orange}{noun} and \textcolor{green!60!black}{body part} are meaningfully distinct in that \textcolor{orange}{noun} is what the subject is touching, and \textcolor{green!60!black}{body part} is the body part they're doing it with. Notably, each of verb, noun, and body part can either come from a predefined list known to be an effective partition for the range of baby touch, or for special cases, an open-ended text input. A touch event can also have multiple annotations for a single field to smooth the decision boundary between similar categories. We provide our 9 predefined verbs, 10 frequently occurring nouns, 5 body parts, and 4 additional conditions in Appendix \ref{app:taxonomy-definitions}. Previously, Lederman and Klatzky~\cite{ledermanHandMovementsWindow1987}  parameterized touch events by the attributes of the object being touched, like shape, size, and weight, and they found that a given child relies on only one attribute to classify objects. Our touch taxonomy aligns with their finding in that the object name subsumes attributes. Moreover, we argue that a touch event's verb, body part, and other sensory conditions are important for describing the whole touch experiences. Altogether, we believe that because of our iterative and data-guided design approach, our taxonomy achieves satisfactory specificity and clarity for describing a baby's touch.

\begin{figure*}
    \centering
    \includegraphics[width=1\linewidth]{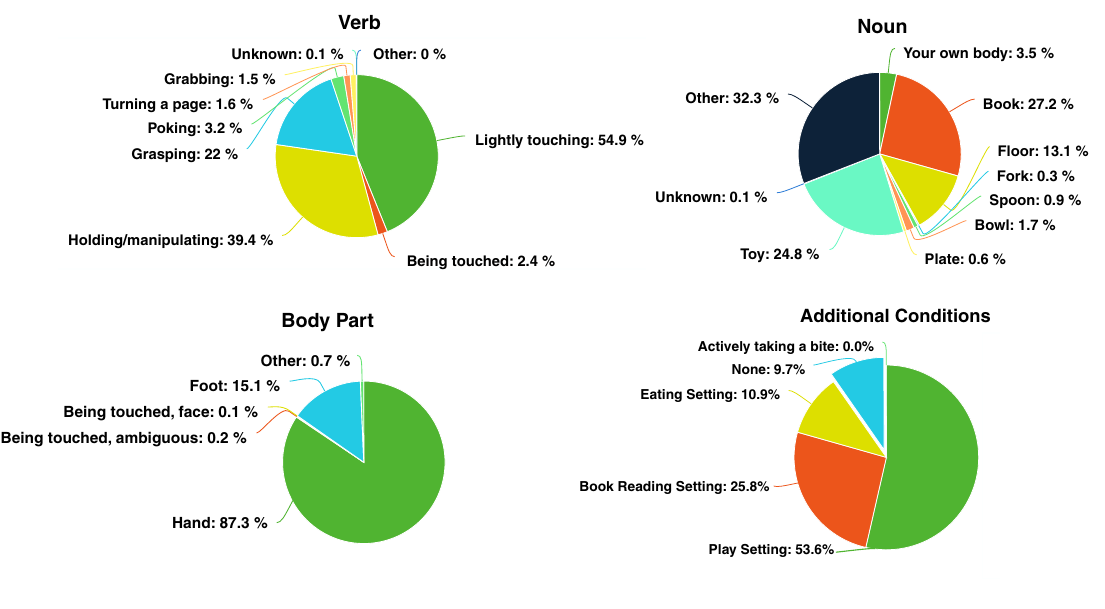}
    \caption{Distributions of the coded touch, [verb, noun, body part, additional conditions], over our touch annotations}
    \label{fig:piecharts}
\end{figure*}

% \boqing{Probably discuss how we came up with this taxonomy, which facilitates our annotation process. There are taxonomies about TOUCH in other contexts, e.g., \url{https://pmc.ncbi.nlm.nih.gov/articles/PMC2991392/}.}\\

\smallskip
\noindent\textbf{Human-coded set:} 
Once we have finalized the taxonomy, we have a set of 2,000 video clips from our final design round, each of which is annotated by 3 human annotators. 858 of these have detected touch events, while the remaining 1142 meet our exclusion criteria. For 79\% of the video clips, the three volunteers unanimously agree to either include or exclude the clip for any reason; most of the remaining 21\% may contain some borderline touch event, but due to the inherently subjective nature of the `Unclear' exclusion criteria one or two of the annotators voted to exclude it. For events detected by the majority of annotators, there is nearly always a consensus on the categorization -- 97\% and 100\% of the time for verbs and body parts, respectively. We construct a human annotated set via majority voting from the three annotators for each clip. As we are satisfied with the expertise of our trained annotators, we consider this set to be the ``ground truth'' labeling policy for generated labeling at scale.

\subsection{Automated coding at scale (Figure~\ref{fig:data-pipeline}, stage 2)} \label{sec:label-at-scale}
The iterations with human raters allowed us to gain sufficient understanding of the touch events and to write clear coding instructions that engaged more raters with no background in our work. To scale up this effort, however, we resort to Gemini~\cite{gemini} coding with humans in the loop. 

\subsubsection{Gemini coding} 
Using the human-coded high-quality set for quality validation, we iterate until finding a highly performing prompt for Gemini to generate touch labels, which we include in Appendix \ref{app:prompt}. With our curated prompt and three evenly sampled frames per video clip, Gemini-3-pro~\cite{gemini} detects touch events with 77\% accuracy, 68\% precision, and 87\% recall relative to the human majority votes. Further, Gemini-3-pro predicts the human-coded verb, noun, and body part with 62\%, 82\%, and 93\% accuracy, respectively. 

%\subsubsection{Caption template} 
%For each structured touch label, we can create a text caption, for example, (Being touched, parent, leg, none) can be encoded as ''being touched by parent on your leg'', or (lightly touching, book, hand) can be encoded as ''lightly touching a book with your hand''. Conveniently, we can mix "fill in the blank" labels seamlessly alongside predefined categories. \\

\subsubsection{Filtering}
To improve the precision of our generated labels, we need an additional quality control component to remove low quality $[\texttt{frame, touch caption}]$ pairs from our training dataset. To do this, we remove all pairs whose pretrained CLIP~\cite{clip} similarity scores are below 0.2, or, because touch is inherently motion-dependent, pretrained BLIP \cite{blip} (a video-based caption similarity model) scores below 0.04. We choose this BLIP threshold to remove approximately the same fraction of frames, approximately 10\%, as did CLIP. %We find that in practice, filtering pretraining data with a similar model architecture to the one we plan to train helps boost data efficiency.
%After filtering, 

\smallskip
\noindent\textbf{Gemini-coded set:} 
After classifying SAYCam at scale, processing those into meaningful codes, and filtering those codes, we are left with 263,604 $[\texttt{image, touch}]$ training examples. Figure~\ref{fig:piecharts} displays some the relative frequencies of each predefined category, with open-ended categories counted as ''other''. Note that these percentages are only calculated over the valid touching events- for instance, 1.6\% Turning a Page should be interpreted as \textit{1.6\% of touch learning events are in the form of turning a page}, rather than \textit{babies spend 1.6\% of their waking hours turning pages}. Likewise, 53.6\% in a ''Play Setting'' means \textit{53.6\% of touch events happen in a Play Setting}, rather than \textit{53.6\% of a baby's total sensory input happens in a Play Setting.} Our touch annotations disproportionally reflect specific conditions by nature, because touch learning in the wild occurs disproportionally under such conditions. We combine these examples with BabyVLM-V2's 768k speech labels, yielding a dataset of 1.03M total captions over 936k unique frames. We can formalize our notation to consider three types of learning events: 96k [\texttt{vision, speech, touch}] \textbf{triplets}, 672k [\texttt{vision, speech}] pairs, and 168k [\texttt{vision, touch}] pairs.
We show illustrative examples of each kind of event in Figure \ref{fig:models}A.

\subsection{Mapping touch captions to IDs (Figure~\ref{fig:data-pipeline}, stage 3)}
%With a large scale set of learning events constructed from generated categorical labels and \cite{wangbabyvlmv22026}, we move on to our next phase: clustering, embedding, and training a ViT from these events. A visualization of our training algorithm is shown in Figure \ref{fig:pipeline}B. \\
% \textbf{The need for clustering of touch captions: }The need for  \\
% \textbf{Touch Indices: }Although text captions are useful for rapid prototyping and inclusion of "fill in the blank" annotations, they contradict our premise of developmental fidelity by providing the model with language supervision that a baby would not have access to. We want a way to encode the touch labels that maintains their uniqueness but does not directly map them to words. To achieve this, we can compute indices which represent each unique type of touch sensation, for example, \{``lightly touching", "spoon", "hand"\} is represented by the index 1, {"manipulating", "spoon", "hand"\} is represented by the index 2, and so on. The indices can then be mapped to learned embeddings and aligned with the vision representation. Note that not every possible combination of labels appears, for example, {"grasping", "anything", "leg"\} never exists.  We end up with 888 unique touch indices, however, these are difficult to learn from as we have no way to include "fill in the blank" labels without adding a new index for each custom caption.
% \boqing{Probably don't describe the touch indices at all as we didn't use them in the exp. What do you think?}\\

We map the resultant ``touch captions'' (e.g., [\textcolor{red}{manipulating}, \textcolor{orange}{spoon}, \textcolor{green!60!black}{hand}, \textcolor{blue}{eating setting}]) to discrete cluster indices by running a vanilla K-means algorithm~\cite{mcqueen1967some} over the embeddings of the touch captions extracted from CLIP's text encoder~\cite{clip}.  Importantly, we will feed the cluster IDs rather than the ``touch captions'' to our computational models. %model to run a vanilla k-means clustering algorithm on our touch captions and assign them to 16, 64, 256, and 1024 clusters. In practice, we find that \todo{N} clusters is the most effective for our learning algorithm.

\smallskip
\noindent \textbf{The need for clustering:} We want to avoid providing additional supervision to our model that a baby wouldn't have access to in their daily life. By converting the touch signals from a form like ``grasping a spoon'' to ``experiencing a touch sensation of type 4'', we move closer to how a real baby might process touch. Note that the cluster formation still accounts for each category of touch and includes ``fill in the blank'' labels in K-means. 

Moreover, for proper experimentation on the role of touch, we need to move the touch signals out of the language space to simulate the property that the spoken word ``book'' is not the same as the touch sensation of ``book''. Specifically, when the model sees a training example with a speech transcription containing ``book'', it should update its understanding of what the word ``book'' means; when it sees an example with a touch cluster associated with books, it should update its understanding of what that touch cluster means. If we did not include a clustering step in our experimental design and instead trained with the touch captions, these important properties would be lost, and we would be less certain about the role touch plays in the learning.

\section{Computational Models}
\begin{figure}
    \centering
    \includegraphics[width=1\linewidth]{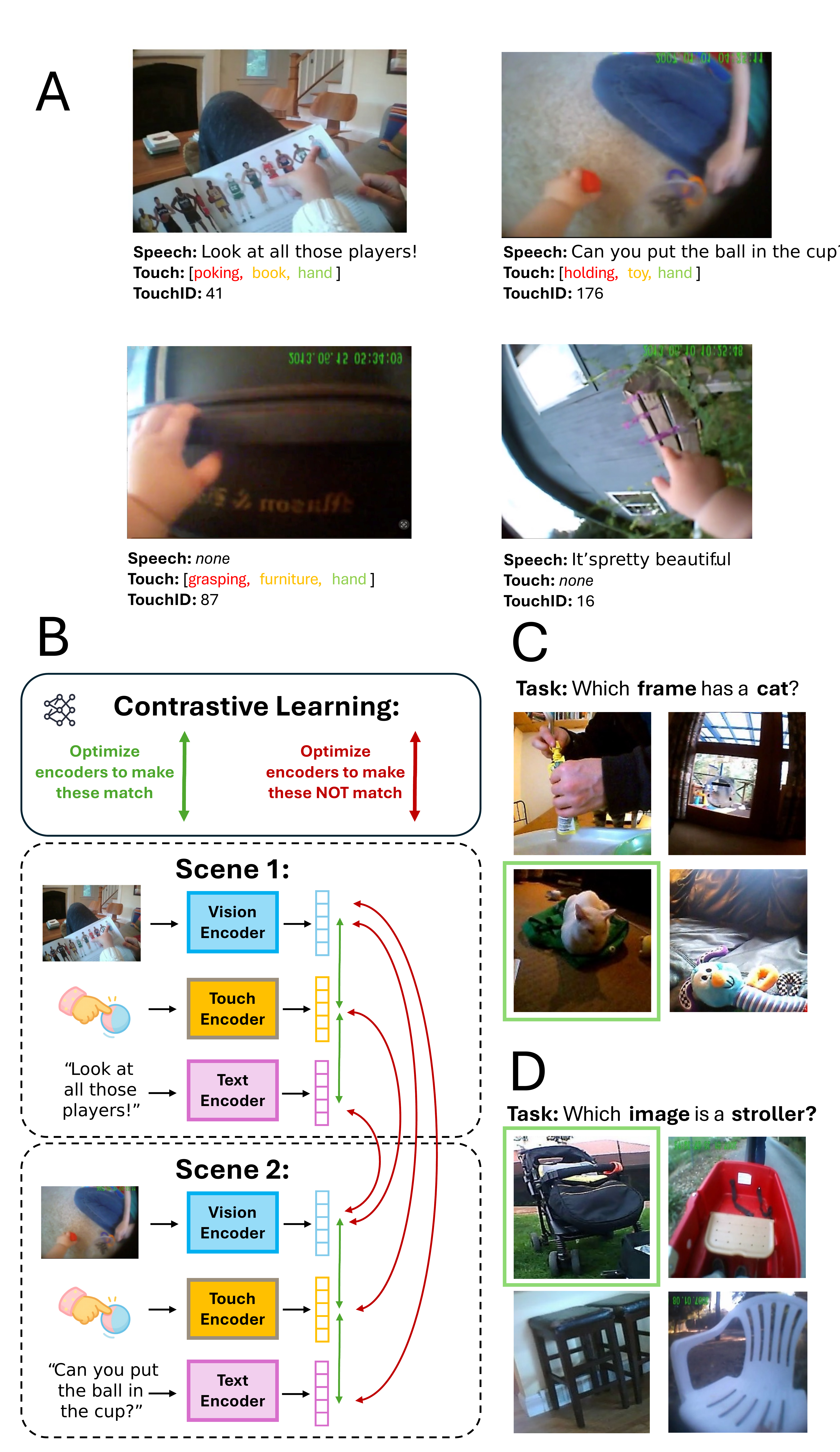}
    \caption{Computational Models. \textbf{A)} Examples of [\texttt{frame, speech transcript, touch ID}] triplets (top) and data pairs missing either speech or touch (bottom). \textbf{B)} Visualization of model architecture and contrastive learning. \textbf{C)} Labeled-S test sample. \textbf{D)} PV test sample.
    }
    \label{fig:models}
\end{figure}

With the resultant datasets of [\texttt{video frame, speech transcript, touch index}] triplets (see examples in Figure~\ref{fig:models}A), we train computation models using a contrastive learning algorithm (see Figure \ref{fig:models}B). We extend CVCL's visual-textual model architecture to three encoders for video frames, speech transcripts, and touch cluster IDs (represented as one-hot vectors), respectively. We then use the video-audio-text transformer~\cite{vatt}'s dual contrastive loss as the training objective, contrasting time-locked [\texttt{frame, transcript}] and [\texttt{frame, touch}] pairs against misaligned pairs. We will release our code for reproducibility. 

We evaluate the resulting model's visual perception on two existing baby learning benchmarks, Labeled-S~\cite{vongGroundedLanguageAcquisition2024} and Picture Vocabulary (PV)~\cite{wangbabyvlmv22026}, both of which task a model with selecting, from four images, the one corresponding to a given category label. The four-choice stimuli images per task in Labeled-S are raw video frames randomly drawn from the same domain as our training dataset, whereas PV is more challenging --- the test stimuli are crops of SAYCam video frames, which makes them slightly outside the training domain; and the distractors are carefully selected to be similar to ground-truth option, either semantically or visually. Samples from both benchmarks are shown in Figure \ref{fig:models}C. For Labeled-S, we report both zero-shot and linear-probing results; the former selects the image with the highest cosine similarity to the given category label, computed over the embeddings of our models, and the latter adds a linear classification layer trained on the training set of either benchmark. For PV we omit zero-shot accuracy because the input images are out of domain.

% \begin{figure}
%     \centering
%     \includegraphics[width=1\linewidth]{figure2.pdf}
%     \caption{Illustration of our full labeling (left) and training (right) pipelines}
%     \label{fig:pipeline}
% \end{figure}

%\subsection{Training setup}
%To answer our questions about the nature of touch learning, we train a ViT on our dataset with a contrastive learning objective (CLIP), which we show in Figure \ref{fig:models}B. Intuitively, a CLIP objective is optimized by using a vision model to encode a batch of frames and an embedding model to encode their corresponding labels, then scoring both models by how close the embedding of each frame is to its own label's embedding, and how different the embedding of each frame is to the embedding of every other label. Then, on each training step we update the weights of the ViT and the label embedding to improve this score, thus contrasting the representations of visual inputs that have different labels. To embed speech captions, we simply learn an embedding for each of the 13k words found in all of \cite{wangbabyvlmv22026}'s utterances. Similarly, we learn an embedding for each precomputed touch cluster. To implement a CLIP loss for a training batch mixed with both audio labels and touch labels, we follow the video-audio-text transformer~\cite{vatt} by computing a CLIP loss over the speech labels and a CLIP loss over the touch labels, and optimizing for the sum of the two terms. Ideally, the resulting vision model should learn to contrast visual inputs from different touch clusters.

% \subsection{Evaluation}

%% file: experiments.tex
\section{Results}

\begin{figure}
    \centering
    \includegraphics[width=0.9\linewidth]{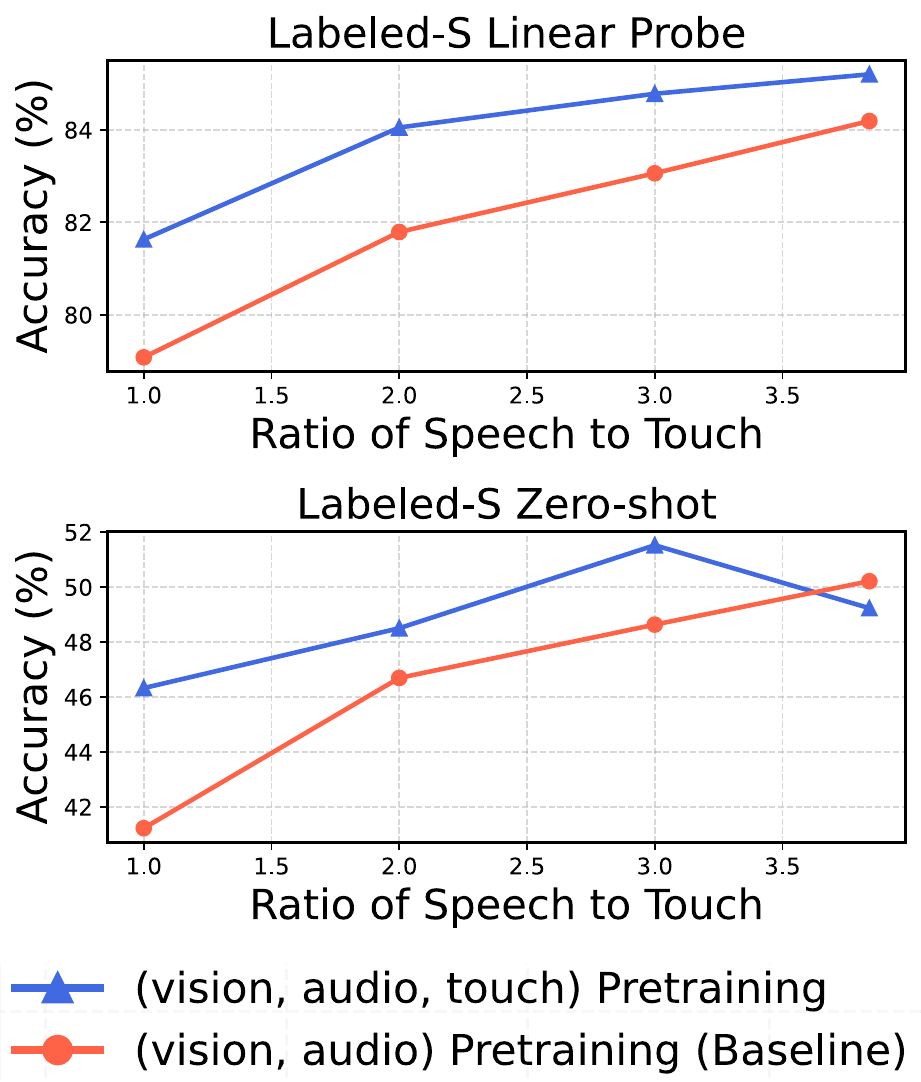}
    \caption{Labeled-S evaluation of the baseline pretraining dataset (red) vs.\ touch-enhanced pretraining dataset (blue), for both linear probing (top) and zero-shot (bottom).}
    \label{fig:plot-labeled-S}
\end{figure}

% \begin{figure}
%     \centering
%     \includegraphics[width=1\linewidth]{Picture Vocabulary_line_plot.pdf}
%     \caption{Main results showing accuracy increase with touch pretraining for Labeled-S and selected tasks from DevCV Toolbox.
%     \label{fig:pv}
% \end{figure}
\begin{figure}
    \centering
    \includegraphics[width=1\linewidth]{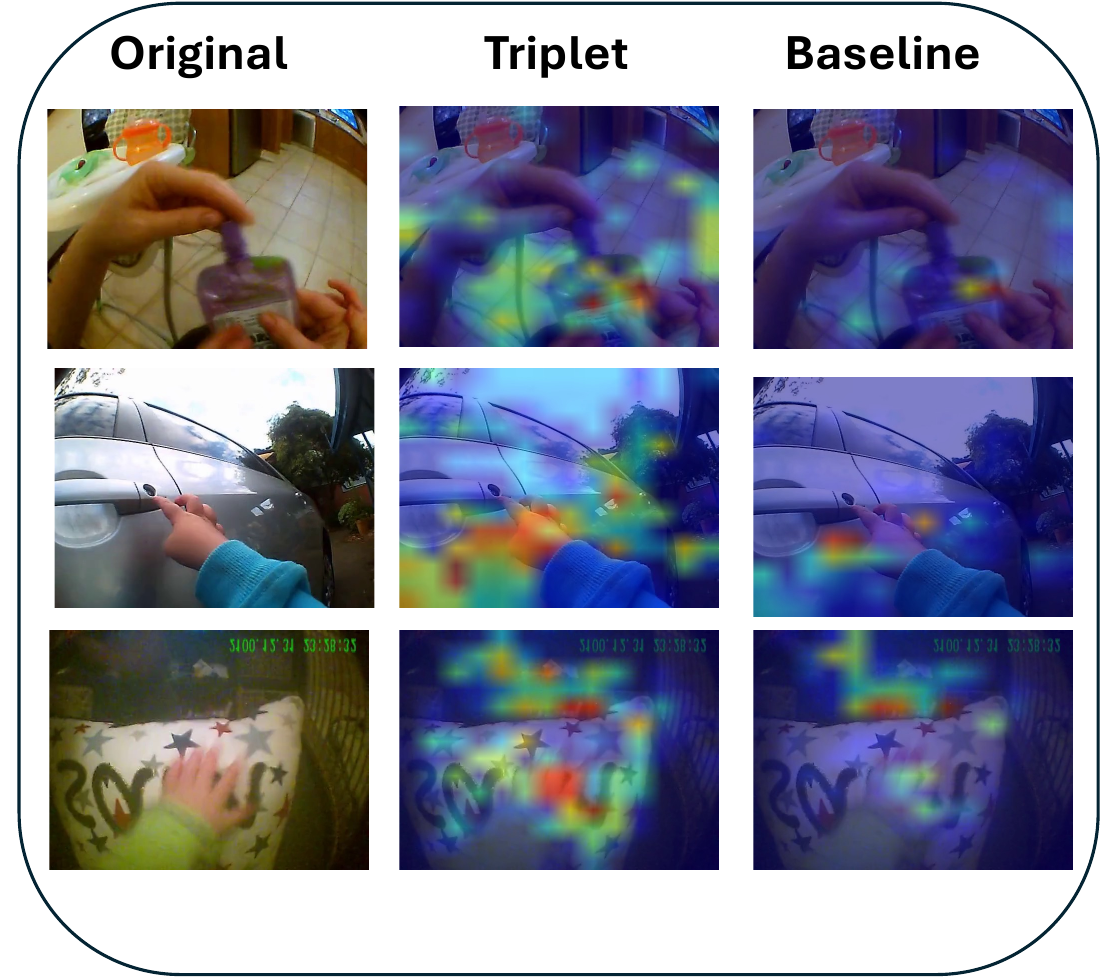}
    \caption{Representative pretraining examples (left), and corresponding attention maps of our model (middle) and a baseline model (right). }
    \label{fig:attention_maps}
\end{figure}
\begin{figure}
    \centering
    \includegraphics[width=0.9\linewidth]{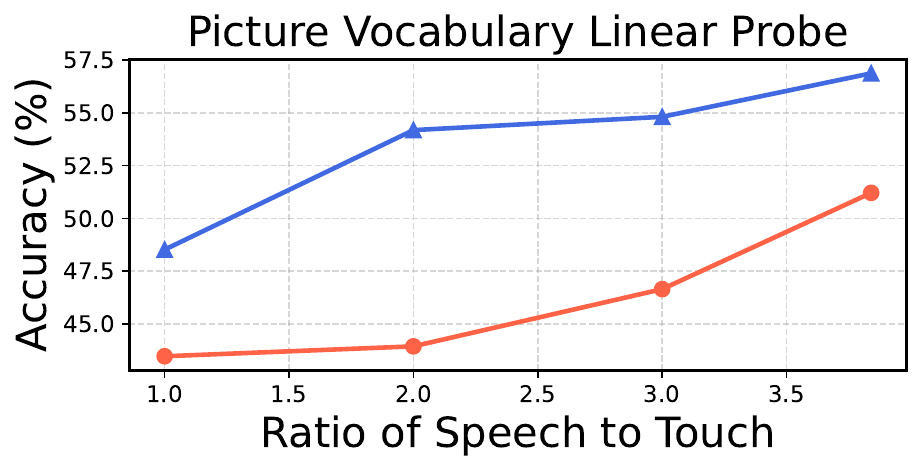}
    \caption{Linear-probing accuracy on PV as a result of the baseline pretraining dataset (red) vs.\ touch-enhanced pretraining dataset (blue)}
    \label{fig:pv}
\end{figure}

We report results that address our overarching research question: Is touch a mechanism for perceptual learning? If it is, how much does our model gain from it relative to other senses? We also conduct a minimal ablation study by reporting the performance of randomly intialized weights. 

\begin{table}
\small
\caption{Linear-probing accuracy on PV and Labeled-S}
\label{tab:linear-probing}
\vspace{-10pt}
\begin{center}
\begin{tabular}{c|cc}
\toprule
\textbf{\%} & \textbf{PV} & \textbf{Labeled-S}\\
\midrule
Random weights (lower bound)&43&65 \\
(vision, speech)&51&84\\
(vision, speech, touch)& \textbf{57}& \textbf{85}\\
\midrule
Random guess & 25 & 25\\
\bottomrule
\end{tabular}
\end{center}
\end{table}

\subsection{Is touch effective as a learning signal?}
Table~\ref{tab:linear-probing} shows the linear probing accuracy for three model variants. We consider:
\begin{itemize}
    \item \textbf{Random weights}: a baseline with no pretraining, using the same visual encoder architecture as the other two model variations with random weights,
    \item \textbf{(vision, speech)}: a contrastive learning model pretrained on our dataset without touch codes, and
    \item \textbf{(vision, speech, touch)}: a contrastive learning model pretrained on our dataset with both speech captions and touch signals discretized into 256 clusters.
\end{itemize}
We also report the chance rate of a random guess.

For both PV and Labeled-S, we see a promising performance boost from touch codes, as shown in Table \ref{tab:linear-probing}. Moreover, Figure~\ref{fig:plot-labeled-S}B reports that for some data ratios, touch gives rise to significant improvement on Labeled-S under the zero-shot setting on Labeled-S. Since the touch learning events are not encoded in the language space, a zero-shot performance boost indicates that the touch pretraining truly shaped the space of vision embeddings, rather than just projecting touch clusters to visual input. 

Moreover, pretrained models on both time-locked datasets significantly outperform the random-weight baseline, echoing Smith and Gasser's proposal that \textit{``multiple overlapping and time-locked sensory systems enable the developing system to educate itself --- without defined external tasks or teachers --- just by perceiving and acting in the world''~\cite{six-lessons}.} 

For a qualitative analysis, as shown in Figure \ref{fig:attention_maps}, we generated attention maps for the \textbf{(vision, speech, touch)} and \textbf{(vision, speech)} models processing pretraining examples. In general, we observe that the model supervised with touch signals learned to favor visual information from regions surrounding hands and the objects they interact with more than the baseline model, qualitatively validating our setup as a computational model of human visual attention.

%We evaluate our model against the one trained only on (vision, speech) pairs. We evaluate the resultant models on Labeled-S, an in-domain object recognition task for which both zero-shot and linear probe results are reasonable, and Picture Vocabulary, an out of domain (OOD) task for which we provide linear probe performance. While Picture Vocabulary is constructed from SAYCam, we consider it to be OOD because the input images are object crops, not frames. For both tasks, we see a promising performance boost when learning from touch labels. Our main results can be seen in Figures \ref{fig:ls} and \ref{fig:pv}. Our observed zero shot improvement on Labeled-S is our most important finding (Figure \ref{fig:ls}); since the touch learning events are not encoded in the language space, a zero shot performance boost indicates that the touch pretraining truly shaped the space of vision embeddings, rather than just projecting touch clusters to visual input. 

\subsection{Where is touch (not) helpful?}
% \boqing{I kept this paragraph for now. Better have the results.} 
We also evaluate the model variations on DevCV Toolbox~\cite{wangbabyvlmv22026}, a computer vision suite developmentally aligned with the NIH Baby Toolbox~\cite{han2025nih}, and find that for very challenging tasks (Counting, WhoHasMore, Localization, Spatial Details) and very simple tasks (Left/Right and Memory), there is no meaningful increase for learning from the touch signals over visual-speech data, suggesting that touch is most useful for the visual-word mapping tasks of Labeled-S and PV. Because of this finding, for our remaining results we focus on Labeled-S and PV, however, we briefly outline our setup and results for the other tasks in Appendix~\ref{evaluation}. Altogether, our results suggest that for certain types of visual reasoning, but not all, touch is helpful in learning.
 % Note that DevBench is intended for evaluation of Vision Language Models (VLMs), while we train only a ViT with task specific linear projections, for which certain tasks (Left/Right, Memory) are less applicable. 
% \begin{figure}
%     \centering
%     \includegraphics[width=1\linewidth]{clustered_stacked_bar.pdf}
%     \caption{\maxwh{should we drop this figure? }Bar chart showing per-task gain achieved by incorporating touch learning examples. For each task and evaluation setup, the salmon-colored box represents the \% gain in accuracy.}
%     \label{fig:barchartperformancegain}
% \end{figure}
\subsection{What is an effective ratio of touch to speech supervision?}
To what extent could touch be helpful in visual learning? Intuitively, the effectiveness of touch sensations would diminish as the visual-speech data increases in size and quality. Figures~\ref{fig:plot-labeled-S} and \ref{fig:pv} speak to this intuition and show how evaluation accuracy scales with the quantity of speech data available during pretraining. Of course, the more speech data available during training, the higher the performance for both Labeled-S and PV, with and without touch events. However, another general trend is that for our in-domain task, Labeled-S, if less speech data is available during training, the effect of touch data tends to be greater, and this holds especially true in the zero-shot setting. Simply put, touch is most important when the speech transcripts are limited --- for a baby with little life experience, it is. Using this insight, we can think of babies as data-efficient learners, integrating as many senses as possible to inform their innate representation of the world~\cite{six-lessons}. 
%While this boost could be attributed to 

% \begin{figure}
%     \centering
%     \includegraphics[width=1\linewidth]{line_plot.pdf}
%     \vspace{-20pt}
%     \caption{As we increase the number of speech training examples, performance on Picture Vocabulary improves.}
%     \vspace{-10pt}
%     \label{fig:ratiolineplot}
% \end{figure}

\subsection{How fine-grained should the touch clusters be to provide effective supervision?}
\label{exp:clusters}
\subsubsection{Hypothesis}
We can use our framework to draw insights into babies' internal representations of touch sensations, specifically, how many distinct types of touch they consider. Intuitively, if we group the touch captions into fewer clusters, we end up treating dissimilar touch sensations as the same and learn a less discriminative representation---it is not necessarily negative because the model can then learn rapidly and data-efficiently to arrive at simpler decision boundaries. If we define more unique clusters, we may consider some similar touch sensations that vary by trivial factors to be different, and the resulting representation can be highly sensitive, implying that a model needs larger-scale data to capture more detailed representations. Tackling real-world visual tasks requires drastically different levels of representation granularity. Hence, it is important for both a baby and a computational model to adaptively leverage the touch sensations depending on specific tasks and situations.

\subsubsection{Labeled-S and PV benefit the most from different levels of touch granularity}
Performance for different numbers of clusters of touch is shown in Table~\ref{clustertable}. Fortunately, our evaluation tasks are well suited to investigate our intuition about the number of clusters: PV quests detailed semantic understanding, as the samples are constructed from crops of semantically similar objects (e.g., sofa and bed), while Labeled-S is more general, like discriminating a video frame with a cat in it from a frame with a window in it. Interestingly, our results confirm our intuition that defining more clusters is important for capturing details, while fewer clusters make it easier to capture general semantics: the model trained with the 256 touch clusters performs the best on PV, and the model trained with 16 clusters performs the best on Labeled-S. %There is no clear answer regarding the optimal number of clusters in general, which we leave to future work. 
% \begin{table}[h]
% \caption{Linear Probe accuracy on selected evaluation tasks for different numbers K of touch clusters}
% \label{clustertable}
% \begin{center}
% \begin{tabular}{c|ccc}
% \hline
% \textbf{K} & \textbf{PV} & \textbf{Labeled-S (linear probe)} & \textbf{Labeled-S (zero-shot)}\\
% \hline
% 16 &54 &\textbf{86}&\textbf{51}\\
% 64 & 57 & 85 & 50  \\
% 256 & \textbf{57}&85&49\\
% \end{tabular}
% \end{center}
% \end{table}

\begin{table}
\small
\caption{Comparing different numbers K of touch clusters}
\label{clustertable}
\vspace{-10pt}
\begin{center}
\begin{tabular}{c|ccc}
\toprule
\textbf{K} & \textbf{PV} & \textbf{Labeled-S (probe)} & \textbf{Labeled-S (zero-shot)}\\
\midrule
\textbf{16}&54 &\textbf{86}&\textbf{51}\\
\textbf{64} & 57 & 85 & 50  \\
\textbf{256}& \textbf{57}&85&49\\
\bottomrule
\end{tabular}
\end{center}
\end{table}

% \subsection{Does the order of learning matter?}
% \maxwh{Highly considering dropping this experiment, I don't think it's very well done.}
% Psychology literature tells us that for the first 8 months of life, humans learn from audiovisual supervision, then upon acquiring this base level of intelligence they learn locomotion and begin to use touch as a learning mechanism. In fact, we find a similar phenomenon when learning visual understanding from speech and touch clusters. When we train on (vision, audio) pairs until convergence then begin to mix in (vision, touch) pairs and (vision, audio, touch) triplets, we see our highest measured intelligence. When we train on all three modalities from the very start, we see lower final intelligence as measured by DevBench.
% \maxwh{have numbers I can provide for this section if we decide to keep it}
% These results can be seen in \todo{small table of this}.
% \subsection{Ablations}
% Finally, to further verify the effectiveness of our touch dataset, we conduct two ablation studies. First, we remove all pretraining data, and measure a randomly initialized model  with only a trained MLP projection's performance on DevBench and Labeled-S. Additionally, following \cite{vongGroundedLanguageAcquisition2024} we assign random touch clusters to the training examples. Results for both ablations are included in Table \ref{tab:linear-probing}.

%% file: conclusions.tex
\section{Conclusions}
We provide a computational framework for modeling touch behavior in babies and demonstrate its effectiveness through structured experiments. Our results quantitatively reveal the effect of touch on babies' acquisition of visual concepts. Still, the exact mechanisms that drive it remain open questions --- is there anything inherently important about the experience of touching, or is it just a means to enriching the visual learning stream? How interrelated is the supervision between time-locked touch events and speech utterances? At what developmental stage do babies start using touch to educate their visual system? 

Moreover, we leave two intriguing hypotheses for future work. One is that the touching moments contain high-quality visual information (in the whole view) for learning, and the other is that the model learns to use and weigh visual information around a touched area for object learning, since the touched object is likely to be labeled in speech. 

Our framework has the potential to serve as an open platform to advance understanding of such mechanisms and accelerate interdisciplinary research between developmental psychology and computer science.

%% file: appendix.tex
\section*{Appendix}

\subsection{Descriptions of touch categories}
\label{app:taxonomy-definitions}
Here we describe how we partition the range of baby touch into predefined verbs in a way that is both specific and consistent. \\
\textbf{Verbs (9)}
\begin{itemize}
    \item \textbf{Lightly Touching:} The most common verb, feeling an object or surface to explore its its texture, temperature, or some other property.
    \item \textbf{Being touched: }Visibly being actively touched by another person.
    \item \textbf{Holding/manipulating: }Lifting, pushing or moving an object like a fork or a ball.
    \item \textbf{Grasping: }Similar to holding/manipulating, but the object is stationary. Also similar to lightly touching, but the hand is wrapped around the object or pushing on it rather than merely touching it. Common for railings, tabletops, etc.
    \item \textbf{Poking: }A subset of lightly touching that occurs very frequently in the baby domain.
    \item \textbf{Turning a page:} A subset of holding/manipulating which also shows up frequently enough to warrant its own category.
    \item \textbf{Grabbing:} The action of initiating a period of holding/manipulating an object.
    \item \textbf{Unknown}
    \item \textbf{Other: }All of the "fill in the blank" verbs go here.
\end{itemize}
\textbf{\textbf{Nouns (10)}
\begin{itemize}
    \item \textbf{Book}
    \item \textbf{Toy}
    \item \textbf{Floor}
    \item \textbf{Your own body}
    \item \textbf{Bowl}
    \item \textbf{Spoon}
    \item \textbf{Plate}
    \item \textbf{Fork}
    \item \textbf{Unknown}
    \item \textbf{Other} (fill in the blank)
\end{itemize}
\textbf{Body Parts (5)}
\begin{itemize}
    \item \textbf{Hand}
    \item \textbf{Foot}
    \item \textbf{(being touched on an) Ambiguous body part}
    \item \textbf{(being touched on the) Face}
    \item \textbf{Other} (fill in the blank)
\end{itemize}}

\textbf{ \textbf{Additional Conditions (5)}
\begin{itemize}
    \item \textbf{Book Reading Setting}
    \item \textbf{Eating Setting}
    \item \textbf{Play Setting}
    \item \textbf{Actively taking a bite}
    \item \textbf{Other} (fill in the blank)
\end{itemize}}

\subsection{Prompt used to query touch captions}
\label{app:prompt}
\textbf{Prompt:}
You are a specialized video annotation assistant. You will be provided with 3 frames from a video clip.

Your task: output ONE valid JSON object that follows the schema below.

Think step-by-step internally to decide the correct labels, but DO NOT output your reasoning. Output ONLY the final JSON.

When deciding, consider these aspects internally (not necessarily in any fixed order):
\begin{itemize}
    \item Perspective: is this clearly the egocentric point-of-view of the INFANT (camera mounted on infant)?
    \item Infant evidence: are INFANT body parts visible (hands/feet/legs/torso edge)? Do NOT treat adult-only body parts as infant evidence.
    \item Contact evidence: is there visible physical contact between an INFANT body part and some object/surface/body? If you cannot confidently confirm infant contact, choose Unclear or No touch as appropriate.
    \item Third-person infant: if you see the infant/toddler in third-person view (the "Blond Toddler Rule"), label Not egocentric.
    \item Ambiguity: if frames are blurry/dark/occluded and you cannot confidently label the infant's interaction, you MUST select Unclear.
\end{itemize}

CRITICAL RULES (must follow):
\begin{itemize}
    \item You are annotating TOUCH sensations/actions of the EGOCENTRIC INFANT ONLY.\begin{itemize}
        \item  NEVER annotate the caregiver's touch/actions.
        \item If the caregiver touches an object but the baby does not, do NOT mark Touch for the baby.
    \end{itemize}
    \item The "Blond Toddler Rule" overrides everything:
    \begin{itemize}
        \item If you see a small infant/toddler in third-person view, set mutually\_exclusive = "Not egocentric".
    \end{itemize}
    \item Egocentric requirement:
    \begin{itemize}
        \item Only label "Touch (default)" or "No touch BY THE BABY" if the camera is clearly infant egocentric POV.
        \item If egocentric POV is not clearly confirmed, prefer "Not egocentric" (if third-person infant visible) else "Unclear".
    \end{itemize}
    \item Use ONLY what is visible in the provided frames. No guessing beyond evidence.
\end{itemize}
OUTPUT FORMAT:
Return ONLY a valid JSON object. No markdown. No extra text. Use exactly this structure and keys:

\lstset{
  breaklines=true,
  breakatwhitespace=false,
  basicstyle=\ttfamily\small,
  columns=fullflexible,
  postbreak=\mbox{\textcolor{gray}{}\space}, % optional break marker
}

\begin{lstlisting}
{
  "mutually_exclusive": "Touch (default) | No touch BY THE BABY | Unclear | Not egocentric",
  "checkbox_q1": ["Book Reading Setting", "Eating Setting", "Play Setting", "ACTIVELY taking a bite"],
  "text_q1": "Other sensory decription",
  "checkbox_q2": ["Lightly Touching (default)", "Being touched", "Holding/manipulating", "Grasping", "Poking", "Turning a page", "Grabbing", "Unknown"],
  "text_q2": "Enter another action not listed",
  "checkbox_q3": ["Your own body", "Book", "Floor", "fork", "spoon", "bowl", "plate", "toy", "unknown"],
  "text_q3": "Another custom object category",
  "checkbox_q4": ["hand (default)", "Being touched on an ambiguous body part", "Being touched on the face", "Foot"],
  "text_q4": "Fill in the blank with some other body part"
}
\end{lstlisting}

DATA TO ANALYZE:
\begin{itemize}
    \item Focus exclusively on the infant's body parts and what THEY touch.
    \item Ignore adult hands/movement unless the baby is touching them (adult can be the noun only if baby touches).
\end{itemize}

\subsection{Evaluation}
\label{evaluation}
We model our evaluation pipeline after \cite{vongGroundedLanguageAcquisition2024}, measuring linear probe accuracy, and zero shot accuracy where applicable. For reproducibility, we now provide brief descriptions of the formats and implementations for each evaluation task. \\
\textbf{Picture Vocabulary and Labeled-S: }
In both of these tasks, the model receives as input four images and a target object name, and is measured on its ability to recognize which image contains the named object. In Picture Vocabulary, all four images are cropped about object bounding boxes, and the incorrect answers are selected to be semantically similar to the target. In Labeled-S, the input images are uncropped SAYCam frames, but the objects found in the input images are unrelated. For zero shot accuracy, we encode each of the images with the pretrained Vision Transformer (ViT), and encode the target word with its pretrained audio embedding. We recognize the image embedding with the highest cosine similarity to the word embedding as the model's answer. For the linear probe, we learn a single linear projection layer on top of the ViT's embeddings and a similar layer on top of the noun embeddings, which gives a reasonable performance boost. For Labeled-S, we predefine an 80/10/10 train/test/val split for all linear probing experiments.\\
\textbf{Localize: }This task evaluates the model's ability to detect which quadrant of a frame a named object appears in. For the linear probe setup, we train a 3-layer MLP probe to output the quadrant the named object is most likely to be in, given the pretrained embeddings of the input frame and the target object's name. We exclude the examples with target objects out of our vocabulary (250 examples of the 2100). Localization is a difficult task to learn past random performance- in our implementation, a randomly initialized ViT and the (vision, speech) baseline model both score 35\% while the model trained on touch scores 38\%, but we omit these results from our interpretation of touch as we consider all of the performance too low to constitute a meaningful learning result. \\
% \textbf{Spatial Details: }Here we evaluate the ViT's ability to encode certain details in images. In the zero-shot setting, we recognize the option image with the highest cosine similarity to the target as the model's answer, the linear probe setting is identical except for a learned linear projection applied to each image embedding. \\
% The memory task takes a reference image and two candidate images as input: one is a repeated instance of the reference image, and the other is a new image. The objective is to identify the new image. In the zero-shot setting, we compute the cosine similarity between the reference image and each candidate, and select the one with lower similarity as the new image; this procedure consistently achieves 100\% accuracy.
% For linear probing, we concatenate the features of all three images and feed them into a residual MLP probe with seven linear layers. The probe is trained to predict the index of the new image. After training, all backbones achieve nearly 100\% accuracy. \\
\textbf{Counting and Who Has More: }Counting evaluates the ViT's ability to determine the quantity of a named object that occurs in a synthetic input image. We evaluated the counting accuracy of a 7 layer residual MLP probe on the target noun and image embeddings of our pretrained models. Our baseline model and touch model both score 35\%; although this is notably above a lower bound (27\%) and below an upper bound (59\%), we do not include it in our experiments as it is not clear that touch learning had a measurable impact. Based on this finding, we focus on Picture Vocabulary and Labeled-S, leaving evaluation of touch datasets on both Count and Who Has More to future work.\\
% \textbf{Who Has More}\\
% This tests a ViT's ability to identify which of two SAYCam frames contains more of a named object. Again, a zero-shot setting is not applicable, so we report accuracy of a 3 layer MLP probe on the two input frames. Based on our Counting\\
\textbf{Memory, Left/Right, Spatial Details and Visual Delayed Response: }Because the DevCV Toolbox was designed to evaluate vision language Q\&A models, three of the tasks, Memory, Left/Right, and Spatial Details, are too simple to provide a meaningful signal on the quality of a ViT due to its architecture. We leave an evaluation of touch models on the only video-based task in the DevCV Toolbox, Visual Delayed Response, for future work as we pretrained with single frames only.   \\

%% file: acknowledgemnts.tex
\section*{ACKNOWLEDGMENT}
Special thanks to Jessica Sullivan and Arjun Chandra for their support and feedback throughout the project.
Additionally, thanks to Aryan Sharma, Batyrkhan Baimukhanov, Raman Deep Shiva Murthy, and Yu Sheng for their help annotating touch events. The project is supported in part by NSF 2540851 and a Sony Faculty Innovation Award. 